\documentclass[10pt,twocolumn,letterpaper]{article}

\usepackage{iccv}
\usepackage{times}
\usepackage{epsfig}
\usepackage{graphicx}
\usepackage{amsmath}
\usepackage{amssymb}
\usepackage{booktabs}
\usepackage{multirow}
\usepackage{caption}
\usepackage[ruled]{algorithm2e}
\usepackage[normalem]{ulem}
\usepackage{makecell}

\newcommand{\mat}[1]{\mathrm{\textbf{#1}}}
\newcommand{\vect}[1]{\mathrm{\textbf{#1}}}
\newcommand{\image}[1]{\mathcal{{#1}}}

\newcommand{\at}[1]{\big\langle #1 \big\rangle}

\DeclareMathAlphabet{\mymathbb}{U}{BOONDOX-ds}{m}{n}

\usepackage[pagebackref=true,breaklinks=true,letterpaper=true,colorlinks,bookmarks=false]{hyperref}

\iccvfinalcopy 

\def\iccvPaperID{9688} 

\ificcvfinal\fi

\title{Test}
\author{A. Tester}

\begin{document}

\title{Tiled Multiplane Images for Practical 3D Photography}

\author{Numair Khan\quad Douglas Lanman \quad Lei Xiao \\Reality Labs Research, Meta}


\twocolumn[{%
\renewcommand\twocolumn[1][]{#1}%
\maketitle
\begin{center}
    \centering
    \captionsetup{type=figure}
     \includegraphics[width=0.95\linewidth, trim={0cm, 11.5cm, 10.5cm, 0cm}, clip]{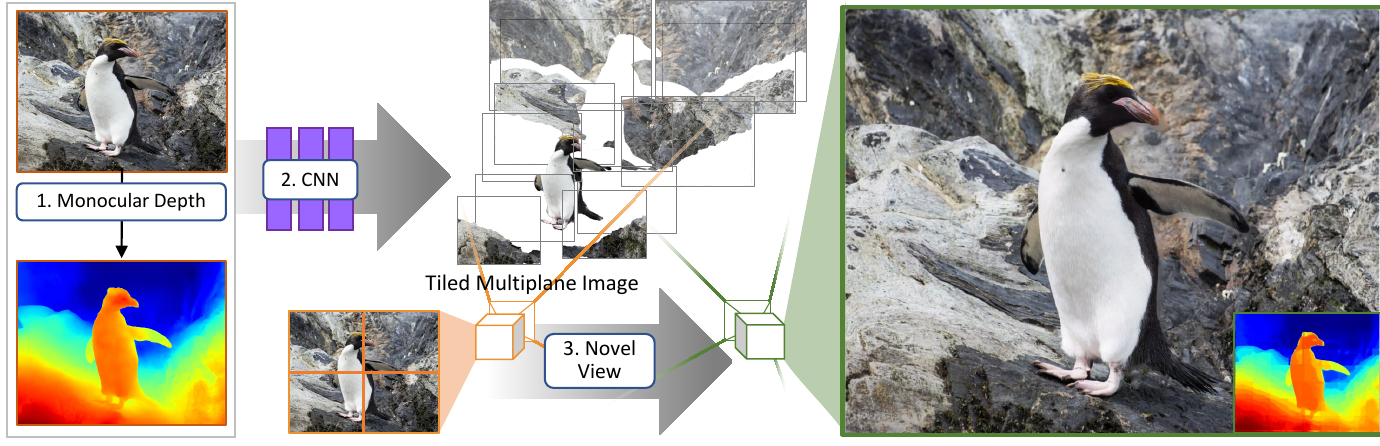}
    \captionof{figure}{An overview of our 3D photography method. \textbf{Left:} Using a single RGB image and an estimated monocular depth map, the scene is recreated as a tiled grid of many small MPIs. \textbf{Middle:} A visualization of a 2$\times$2 tiled grid of MPIs, each with three RGBA layers. \textbf{Right:} Novel views are rendered by warping and compositing each MPI tile into the target camera's frustum. The 768$\times$1152 pixel image shown is generated using a 7$\times$11 grid of 4-layer MPIs. }
\end{center}%
}]

\maketitle
\ificcvfinal\thispagestyle{empty}\fi


\begin{abstract}
The task of synthesizing novel views from a single image has useful applications in virtual reality and mobile computing, and a number of approaches to the problem have been proposed in recent years. A Multiplane Image (MPI) estimates the scene as a stack of RGBA layers, and can model complex appearance effects, anti-alias depth errors and synthesize soft edges better than methods that use textured meshes or layered depth images. And unlike neural radiance fields, an MPI can be efficiently rendered on graphics hardware. However, MPIs are highly redundant and require a large number of depth layers to achieve plausible results. Based on the observation that the depth complexity in local image regions is lower than that over the entire image, we split an MPI into many small, tiled regions, each with only a few depth planes. We call this representation a {\normalfont Tiled Multiplane Image} (TMPI). We propose a method for generating a TMPI with adaptive depth planes for single-view 3D photography in the wild. Our synthesized results are comparable to state-of-the-art single-view MPI methods while having lower computational overhead. 
\end{abstract}


\section{Introduction}
\label{sec:introduction}
The novel view synthesis (NVS) problem involves using a set of input images to generate views from new and unseen camera positions, allowing three dimensional interaction with photos. This is a long-studied problem, with early work relying on interpolation within dense structured image sets~\cite{levoy1996, gortler1996, davis2012}. The specialized rigs commonly required to capture the large number of images restricted these methods to lab settings~\cite{wilburn2005}. However, the potential applications offered by novel view synthesis on modern mobile and VR devices has kindled wide interest in the problem, and encouraged researchers to seek methods that make the technology more accessible. The term \emph{3D photography} refers to the use of novel view synthesis in everyday capture settings, often from a single image. 

Over the past few years a number of proposed scene representations have leveraged the great strides being made in learning-based techniques to achieve more accurate synthesis with fewer constraints. The most recent of these are neural radiance fields (NeRFs) ~\cite{mildenhall2019, xie2022} which represent the scene as multi-layer perceptrons. Their results define the high bar of novel view synthesis. However, this high quality has a significant data and computational cost. 

An alternate  representation, a multiplane image (MPI), defines the scene as a stack of fronto-parallel RGBA planes that can be warped and rendered into novel viewpoints~\cite{zhou2018, mildenhall2019, flynn2019}. An MPI offers the advantage of rendering speed, and suffers from less aliasing than mesh or point-based methods~\cite{shih20, wiles2020, niklaus2019}. The latter characteristic is important for applications
that require temporal stability. However, an MPI is a highly redundant scene representation: the number of RGBA planes required to capture and reconstruct all the depth variation in a scene can be quite high. Since most scenes have a larger amount of free space than occupied, most of the planes in an MPI are very sparse. This makes them inefficient as a representation~\cite{barlow1961}, and expensive to generate, transmit, and store. 

In this paper we propose to address these shortcomings of multiplane images and develop a lightweight solution to the 3D photography problem that can be practically implemented on mobile and VR devices. Some examples of the applications we envision are 3D video conferencing, telepresence, and  VR passthrough~\cite{xiao2022}. We show how subdividing the image plane into many small MPIs with only a few planes in each, provides a more efficient representation from a computational and memory perspective. However, the naive approach of directly using existing MPI methods with tiles creates boundary artifacts in the novel views. This happens because the commonly used fixed spacing of MPI planes fails to capture the full depth range of a tile when the number of planes is small. Furthermore, it is sensitive to outliers in small regions. We propose a clustering-based approach using learnt confidence weights to predict per-tile MPI planes that better represents local depth features. Our method is lightweight and generates results comparable to the state-of-the-art in MPI-based 3D photography. 

\begin{figure*}[t]
   \includegraphics[width=1.0\textwidth, trim={0cm, 13.2cm, 15.5cm, 0cm}, clip]{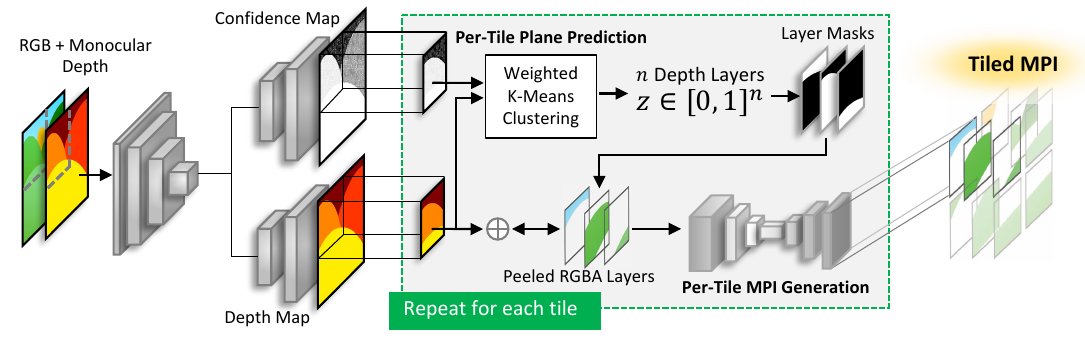}
\caption{Tiled multiplane image generation for single-view 3D photography. Given an RGB image and depth from a monocular estimator, our method first generates a pixel-wise confidence map and a pre-processed depth map. These are used to predict $n$ depth planes per tile through weighted $k$-means clustering (with $k=n$). The layer masks defined by pixel labels are used to peel RGBA layers, yielding a rough initial MPI. This is concatenated ($\bigoplus$) to the depth and passed to a refinement network that generates the RGBA images of the final per-tile MPI. }
\label{fig:pipeline}
\end{figure*}

In summary, the main contribution of this work are,
\begin{enumerate}
    \item The demonstration of \emph{tiled multiplane images} as a practical representation for view synthesis tasks.
    \item A learning framework for generating tiled multiplane images from a single RGB input for 3D photography.
    \item A novel approach to adaptive MPI plane positioning. 
\end{enumerate}

\section{Related Work}
\label{sec:related-work}
The progenitors of current 3D photography were the early works on image-based rendering (IBR). These methods usually relied on interpolation within the convex hull of a large set of images to generate novel views. Levoy~\etal\cite{levoy1996} and Gortler~\etal~\cite{gortler1996} proposed the canonical two-plane parameterization of light fields that renders novel views by quadrilinear interpolation. Gortler's method also provided an early demonstration of the use of geometric proxies to improve rendering quality. Davis~\etal's~\cite{davis2012} work extended interpolation-based view synthesis to unstructured images. The  excellent analysis of plenoptic sampling done by Chai~\etal\cite{chai2000} proved, however, that for large distances the number of images required for view synthesis by interpolation was impractically high. Consequently, the large majority of recent view synthesis methods have relied on learnt priors to overcome the high sampling requirements. 

One of the corollaries of Chai~\etal's analysis was that the sampling requirements for view-interpolation are inversely related to the geometric information of the scene. Thus, many subsequent methods have relied on coarse geometric proxies to improve view synthesis quality~\cite{riegler2020, riegler2021, hu2021}. In Mildenhall~\etal's~\cite{mildenhall2019} method this proxy takes the form of a multiplane image (MPI), which they use to achieve interpolation-based view synthesis that overcomes Chai~\etal's sampling limits. MPIs were first proposed by Zhou~\etal~\cite{zhou2018} who used them for extrapolating novel views outside the convex hull of the input stereo cameras (Tucker and Snavely~\cite{tucker2020} observe that an MPI can be considered as an instance of Szeliski and Golland's~\cite{szeliski1998} earlier ``stack of acetates'' volumetric model). Mildenhall~\etal~\cite{mildenhall2019} and Srinivasan~\etal~\cite{srinivasan2019}, respectively, provide a theoretical analysis of the limits of view interpolation and extrapolation using MPIs. Flynn~\etal~\cite{flynn2019} use learnt gradient descent to generate MPIs from multiple views. Attal~\etal~\cite{attal2020} and Broxton~\etal~\cite{broxton2020} extend the MPI concept to concentric RGBA spheres that can be used for view synthesis in 360 degrees. Wizadwongsa~\etal~\cite{wizadwongsa2021} and Li~\etal~\cite{li2021} replace the discrete RGBA planes of MPIs with continuous neural surfaces to achieve higher quality results. Tucker and Snavely~\cite{tucker2020} use a scale-invariant method that allows them to learn strong data priors that can generate MPIs from a single view. Their approach is in a line of recent work dubbed \textit{3D photography} that aims for novel views from in-the-wild, single-view images. Li~\etal~\cite{li2020}, Han~\etal~\cite{han2022}, and Luvizon~\etal's~\cite{luvizon2021} methods are MPI-based examples of this approach. An MPI, however, is an over-parameterized scene representation. Recent methods~\cite{li2019, ghosh2021, luvizon2021, han2022} have sought to overcome this shortcoming to some extent through a more judicious placement of depth planes. Nonetheless, their high level of redundancy dilates their memory and computational footprint, and limits their wider adoption in mobile and AR/VR applications. 

For such use cases, a more efficient approach to novel view synthesis involves depth-based warping ~\cite{chaurasia2020, xiao2022}, often followed by inpainting~\cite{kopf2019, kopf2020}. Shih~\etal~\cite{shih20} propose to guide inpainting in disoccluded regions of the warped view using a layered depth representation. Li~\etal~\cite{li2022} apply this general approach to 360-degree input. Wiles~\etal~\cite{wiles2020} use a depth map to generate a point cloud of neural features which can be projected and rendered in novel views using a generative network. Choi~\etal~\cite{choi2019} estimate a probability volume instead of a single depth map to handle uncertainty in difficult regions. Niklaus~\etal~\cite{niklaus2019} use segmentation to remove the geometric and semantic distortions from depth that often impair the rendering quality of this approach. Nonetheless, depth-based warping suffers from hard boundaries and is over-sensitive to errors in the depth estimate. While Jampani~\etal~\cite{jampani2021} propose soft layering with alpha mattes to address this shortcoming, MPIs are inherently capable of handling such artifacts via blending.

Finally, our review of view synthesis would not be complete without mentioning neural radiance fields (NeRFs) ~\cite{mildenhall2021, xie2022} which have recently burgeoned in popularity. While great strides are being made in improving the time and data requirements of NeRFs~\cite{yu2021, muller2022}, they remain expensive for interactive applications. Some recent work~\cite{reiser2021, tancik2022, turki2022} shows that decomposing a single large radiance field into smaller sub-components can improve efficiency. This is similar to the approach we adopt for multiplane images.

\section{Method}
\label{sec:method}

A traditional MPI represents the scene as a set of $N$ fronto-parallel planes in the camera frustum of a reference view $\image{I}$. Each plane is associated with an RGBA image. While it is possible to place the planes at any depth, they are usually arranged linearly in disparity (inverse depth)~\cite{zhou2018, flynn2019, mildenhall2019, tucker2020}. A novel view $\image{I}_t$ is rendered by warping the planes into the target camera's image space via a homography, and compositing them front-to-back using the \textit{over}~\cite{porter1984} operator: 
\begin{align}
    \image{I}_t = \sum_{i=1}^N \big( \alpha_i c_i \prod_{j=i+1}^N (1 - \alpha_i) \big)
    \label{eqn:mpi-rendering}
\end{align}
where $\alpha_i$ and $c_i$ are the warped alpha and color channels, respectively, of the $i^{th}$ plane. 

\begin{figure}[t]
\centering
    \includegraphics[width=1.0\columnwidth, trim={0cm, 8.5cm, 11cm, 0cm}, clip]{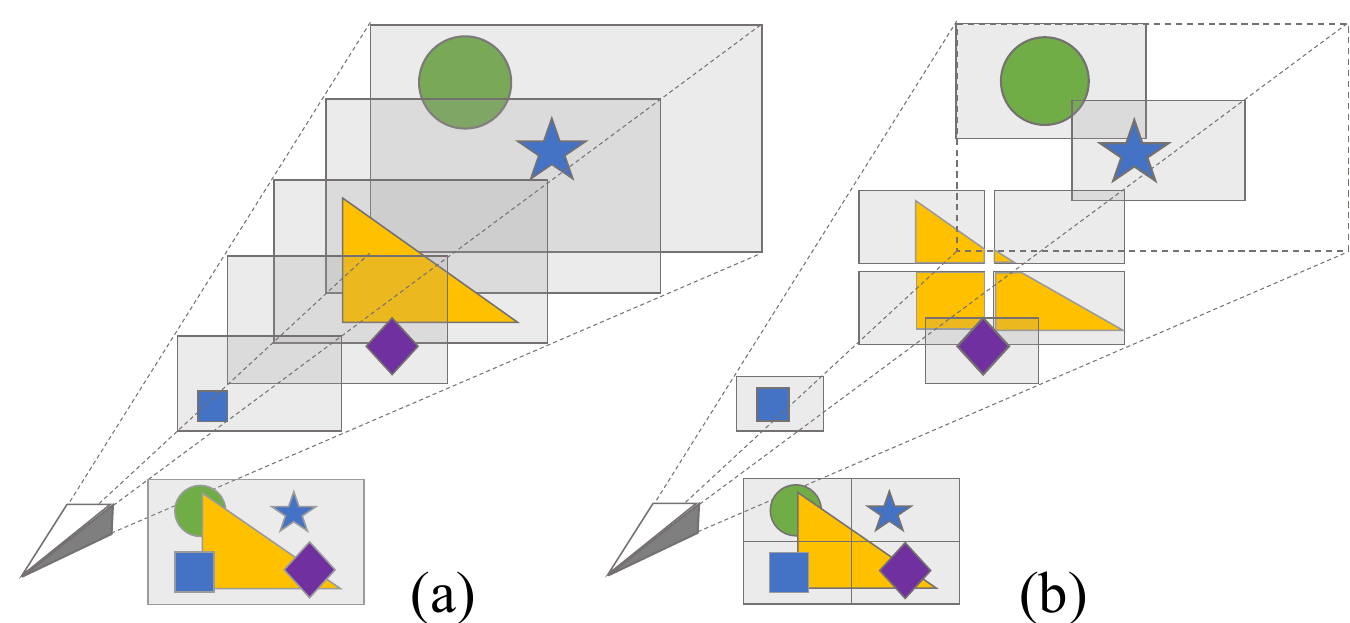}
    \caption{\textbf{(a)} A traditional MPI uses five planes per pixel to represent this toy scene, even though no region has more than two overlapping objects.  \textbf{(b)} A TMPI exploits the low local depth complexity and has only two planes per pixel. }
    \label{fig:tmpi}
\end{figure}

Both warping and compositing can be done very efficiently on graphics hardware, allowing real-time rendering of novel views~\cite{mildenhall2019}. Moreover, the alpha channel at each plane allows MPIs to represent soft edges and anti-alias any errors in the scene reconstruction, leading to fewer perceptually objectionable artifacts than depth-based warping methods. However, the number of planes $N$ required to capture all the depth variation in a scene is usually large, even though most planes are very sparse. We propose to overcome this shortcoming by representing the reference image $\image{I}$ as a tiled grid of many small MPIs~(Figure~\ref{fig:tmpi}). Given $\image{I}$ and its depth map from a monocular depth estimator, our method predicts the placement of $n\ll N$ depth planes within each tiled region and uses this prediction to generate the RGBA images of the MPI in a single forward pass.

\subsection{Tiled Multiplane Image Representation}
\label{sec:tmpi-representation}

Our scene representation is based on a set of $m$ tiles, each representing a square sliding block of size $h$ at 2D pixel locations $\vect{x}_1$, $\vect{x}_2$, ..., $\vect{x}_m$ in the source image $\image{I}$. The locations $\vect{x}_k$ lie on a regular grid with spacing determined by some stride $r$. Each tile consists of $n$ front-parallel RGBA planes, the depth placement of which is not fixed but varies across tiles. We let $\alpha_j^i$, $c_j^i$ and $d_j^i$ denote, respectively, the alpha channel, the color channel, and depth of the $i$-th plane in the $j$-th tile. Then the \textit{tiled multiplane image} (TMPI) representation $\Gamma(\image{I}$) of the image is defined as:
\begin{align}
    \Gamma(\image{I}) = \big\{ (\alpha_j^i, c_j^i, d_j^i, \vect{x}_j) \big\}
\end{align}
for all $i=1, 2, ..., n$ and $j=1, 2, ..., m$. Further, we define an ordering on this set of 4-tuplets as,
\begin{align}
     \big(\Gamma (\image{I}), \leq \big) = \big\{ (\alpha, c, d, \vect{x})_{k=1, ...,mn}~|~ d_k \leq d_{k+1} \big\}
\end{align}
As with traditional MPIs, novel views are rendered in a differentiable manner by warping all planes into the target camera's image space. However, as the depth of the planes varies across the tiles and, hence, across the pixels of $\image{I}$, a planar inverse warp on the target image plane cannot be directly computed via a homography. Instead, each plane must be warped in \textit{tile space} via a homography computed using a shifted intrinsic matrix, and all the warped planes composited sequentially in the target view at their respective tile locations (Algorithm~\ref{algo:render-tmpi}). While this makes the rendering of tiled multiplane images less efficient and somewhat less elegant than MPIs (Equation~\ref{eqn:mpi-rendering}), this is only true during training when differentiability is required. At inference, the tiles can be rendered  as textured quads using hardware-accelerated rasterization making this stage of the pipeline as efficient as traditional MPIs and with lower texture memory requirements. Thus, their compact form makes TMPIs well-suited for rendering over networks, or on mobile devices and VR headsets. The blending of many MPIs locally is related to Mildenhall~\etal's~\cite{mildenhall2019} light field fusion. However, their method renders each MPI separately before blending the results with scalar weights. Our method operates at a much finer scale, and composites the planes of all MPIs together to synthesize a novel view.

\begin{algorithm}[t]
\SetAlgoLined
 \SetKwInOut{Input}{Input}
 \SetKwInOut{Output}{Output}
 \SetKwProg{RenderTMPI}{RenderTMPI}{}{end}
 
 \RenderTMPI{$(\big(\Gamma(\image{I}), \leq \big), \mat{R}, \mat{t}, \mat{K})$} {
  \Input{$\big(\Gamma(\image{I}), \leq \big)$: ordered TMPI planes\\
  $\mat{R}$: relative rotation of novel view\\
  $\vect{t}$: relative translation\\
  $\mat{K}$: camera intrinsics}
  \Output{Novel view $\image{I}_t\in \mathbb{R}^{3\times H \times W}$}
  \vspace{1mm}
  $\image{I}_t \gets \mymathbb{0}^{3\times H \times W}$; $\image{T}_t \gets \mymathbb{1}^{H \times W}$;\\
  \ForEach{$(\alpha, c, d, \vect{x}) \in \big(\Gamma(\image{I}), \leq \big)$}{%
    $\mat{\^K} \gets \mat{K} - \begin{bmatrix} \mat{I} & \vect{x} \\ 0 & 1\end{bmatrix}$; \\
    \ForEach{$\vect{u}\in [1, ..., s]\times[1, ..., s]$}{%
        \vspace{1mm}
        $\vect{n} \gets [0, 0, 1]^T$; \\
        \vspace{1mm}
        $\begin{bmatrix} \vect{u}_s \\ 1 \end{bmatrix} \gets \mat{\^K}\big(\mat{R} - \vect{t}\vect{n}^T/d \big) \mat{\^K}^{-1} \begin{bmatrix} \vect{u} \\ 1 \end{bmatrix}$; \\
        \vspace{1mm}
        $w \gets \alpha \at{\vect{u}_s}~\image{T}_t\at{\vect{u} + \vect{x}}$; \\
        $\image{I}_t \at{\vect{u} + \vect{x}} \gets \image{I}_t\at{\vect{u} + \vect{x}} + w~c\at{\vect{u}_s}$; \\
        $\image{T}_t\at{\vect{u} + \vect{x}} \gets \image{T}_t\at{\vect{u} + \vect{x}} \big( 1 - \alpha\at{\vect{u}_s} \big); $
        
    }
  }

    \KwRet{$\image{I}_t$}
 }
\caption{Differentiable view synthesis using TMPIs. Angled brackets denote pixel indexing. }
\label{algo:render-tmpi}
\end{algorithm}


\subsection{Single-View 3D Photography}
\label{sec:threedee-photography}
We now describe our approach to generating tiled multiplane images from a single RGB input (Figure~\ref{fig:pipeline}). Broadly, our method splits the image plane into a regular tiled grid of learnt confidence and outlier-corrected depth. For each tile, the placement of $n$ fronto-parallel depth planes is determined by clustering pixel depth values weighted by the predicted confidence estimates. This latter step is motivated by the fact that with a small depth plane budget $n$, the commonly used equal spacing in disparity~\cite{zhou2018, flynn2019, mildenhall2019, tucker2020} is wasteful. Thus, the goal is to predict the planes that optimally represent all depth variation within a tile. Using the predicted planes, a fully convolutional network generates the $n$ RGBA images that constitute a per-tile MPI. Unlike Han~\etal's~\cite{han2022} adaptive plane method, we generate the RGBA images in a single forward pass. The resulting TMPI is rendered as a set of textured quads using the rasterization pipeline of graphics hardware. 

In more detail, given the source image $\image{I}$, we first obtain a depth map $\image{Z}$ for it using a monocular depth estimator. A two-headed U-Net $\Theta(\cdot)$ then predicts a confidence map $\image{C}$ along with denoised depth $\image{D}$. The goal is to learn a representation that ameliorates the  sensitivity of the subsequent $k$-means clustering step to outliers. The joint prediction of confidence and depth is similar to the depth-routing of Weder~\etal \cite{weder2020} and the aleotoric uncertainty estimation of ~\cite{kendall2017}. The predicted depth and confidence, and the original color image are then unfolded into a set of $m$ square sliding  blocks of size $h$ and stride $r$: $\{(\image{D}^i, \image{C}^i, \image{I}^i)_{i=1, ..., m}\}$. 

Running $\Theta(\cdot)$ on $\image{I}$ and $\image{Z}$ rather than individual tiles allows it to consider non-local features and avoid undesirable tiling artifacts. Additionally, setting $r < h$ in the unfolding step allows neighboring tiles to overlap. This prevents gaps along tile boundaries and also regularizes per-tile operations across neighbors. However, it also increases the total number of tiles and, thus, the computational requirements. We empirically determine a stride value of $r=h - h/8$ for a good balance between quality and computational efficiency.

\textbf{Per-Tile Planes Prediction:} Next, we predict the $n$ depth planes $\{ z_{j=1, ..., n}^i \}$ that optimally represent the features of the $i$-th tile. The common approach of spacing the planes linearly in disparity grows inaccurate as $n$ becomes small.  Luvizon~\etal~\cite{luvizon2021} place the planes at depth discontinuities identified via the histogram of depth values. However, their approach is sensitive to parameter settings and fails for smooth surfaces which have no discontinuities.  A learning-based approach is adopted by Han~\etal~\cite{han2022} and Li~\etal~\cite{li2020}. The former use multi-headed self-attention to adjust a linear placement. While capable of modeling inter-plane interactions, their method is computationally expensive~(Table~\ref{table:efficiency-comparison}). The latter uses a CNN to directly predict the planes. However, we found that without strong regularization a direct approach lacks topological order and has a strong bias towards a fixed placement. An adversarial loss helps improve this but makes the training more unstable. 

We observe that as depth is known, plane positioning can be posed as a simple clustering problem. Thus, we predict $\{ z_{j}^i \}$ using $k$-means clustering on the depth in each tile $\image{D}^i$.  

Standard $k$-means is sensitive to outliers and can generate significantly different plane predictions across neighboring tiles causing artifacts in novel views. Advantageously, along with the $n$ depth planes, clustering also assigns a label to each input pixel, thereby generating a label map that represents discretized depth. Furthermore, the cluster centers of $k$-means are differentiable with respect to the input samples. Thus, we address the outlier problem by training $\Theta(\cdot)$ to filter the input through a self-supervised reconstruction loss on the discretized depth map generated by a weighted $k$-means. In weighted $k$-means, the cluster centers are updated each iteration using the confidence-weighted mean of the constituent samples. Since we do not directly supervise the depth output of $\Theta(\cdot)$, the network can go beyond outlier-filtering to learn any modifications that improves the discrete reconstruction, and consequently optimizes the placement of the $n$ depth planes within each tile.

\textbf{Per-Tile MPI Generation:} Given $\{z_j^i\}$ and the discrete depth map, we estimate a preliminary MPI per tile by peeling RGBA layers from $\image{I}^i$ using the discrete labels as an alpha mask. The masked RGB regions of each plane are inpainted by upsampling valid values from a Gaussian pyramid. A second network $\Psi(\cdot)$ then refines these estimates to generate the final $n$ RGBA images of the MPI for each tile. Following Zhou~\etal~\cite{zhou2018} and Tucker and Snavely~\cite{tucker2020}, we represent the RGBA output as a pixel-wise blend of the input image $\image{I}^i$ and a learnt background. However, unlike these works we predict a background image $\image{B}^i_j$ per plane:
\vspace{-3mm}
\begin{align}
    \image{W}^i_j &= \prod_{k>j}(1 - \alpha^i_k), \\
    \image{I}_j^i &= \image{W}^i_j\image{I}^i + (1 - \image{W}^i_j)\image{B}^i_j
    \label{eqn:bg-blending}
\end{align}
Where $\alpha^i_k$ is the predicted alpha value for each plane.

\section{Training Procedure}
\label{sec:training}

\begin{table}
\begin{center}
\begin{tabular}{lcccc}
\toprule
\multicolumn{5}{c}{Spaces Dataset} \\
\midrule
Method & PSNR$\uparrow$ & SSIM$\uparrow$ & LPIPS$\downarrow$ & L1$\downarrow$ \\
\midrule
SVMPI~\cite{tucker2020} & 25.42 & 0.748 & 0.210 & 0.040 \\
VMPI~\cite{li2020} & 22.37 & 0.636 & 0.268 & 0.057 \\
MINE~\cite{li2021} & 24.02 & 0.702 & 0.229 & 0.048 \\
AdaMPI~\cite{han2022} & {26.17} & 0.703 & 0.229 & 0.047 \\
Ours & 24.93 & {0.750} & {0.175} & {0.037} \\
\midrule
\multicolumn{5}{c}{Tanks \& Temples Dataset} \\
\midrule
Method & PSNR$\uparrow$ & SSIM$\uparrow$ & LPIPS$\downarrow$ & L1$\downarrow$ \\
\midrule
SVMPI & 17.85 & 0.530 & 0.370 & 0.082\\
VMPI & 16.32 & 0.463 & 0.395 & 0.103\\
MINE & 17.23 & 0.506 & 0.366 & 0.088 \\
AdaMPI & 18.62 & 0.565 & 0.270 & {0.073} \\
Ours & {18.69} & {0.569} & {0.267} & {0.073} \\
\bottomrule
\end{tabular}
\end{center}
\caption{Quantitative evaluation of view synthesis results on the \emph{Spaces} and \emph{Tanks and Temples} multi-view datasets. SVMPI, MINE and AdaMPI use 32 MPI depth planes; VMPI uses 8; our approach uses 4 planes per image tile. }
\label{table:mpi-results}
\end{table}
We train $\Theta(\cdot)$ in a self-supervised manner by minimizing the L1 loss between the input depth $\image{Z}$ and the folded discrete depth maps produced per tile by weighted $k$-means clustering. Then, we freeze $\Theta(\cdot)$ and train the MPI generation network $\Psi(\cdot)$  on a novel view synthesis task by following Han~\etal's~\cite{han2022} warp-back strategy to generate pseudo ground truth multi-view training data. This involves warping single-view images into a target camera using monocular depth, and inpainting disocclusion holes with a specially trained network. The view-synthesis training objective is a combination of VGG, structural similarity ~\cite{wang2004}, and L1 losses on the synthesized color image $\image{I}_t$, weighed as 0.1, 0.25, and 1.0 respectively. For both networks, we use the 111K images of the COCO dataset~\cite{caesar2018} and the monocular depth method of Ranftl~\etal~\cite{ranftl2021}.

\section{Experiments}
\label{sec:experiments}

\begin{table}[ht]
\begin{center}
\setlength{\tabcolsep}{2pt}
\resizebox{\columnwidth}{!}{%
\begin{tabular}{lccccc}
\toprule
 & \makecell{{Params.} \\[-0.1cm]{(M)}}$\downarrow$ &
 GMAC$\downarrow$ &  \makecell{{Runtime} \\[-0.1cm] {(ms)}} $\downarrow$ & \makecell{{Peak} \\[-0.1cm] {(GB)}}$\downarrow$ & \makecell{{Space} \\[-0.1cm] { (MB)}}$\downarrow$\\
\midrule
\small{DPT$^*$} & 123.0 & 110 & 31 & 4.39 & -- \\
\midrule
\small{SVMPI} & 43.5 & 58.0 & 111 & 4.53 & 26.9\\
\small{MINE} & 38.1 & 250 & 110 & 4.91 & 107 \\
\small{VMPI}$^\dag$ & 4.31 & 52.3 & 96.0 & 4.57 & 13.5\\
\small{AdaMPI}$^\dag$ & 19.0 & 288 & 350 & 5.94 & 107\\
\small{Ours}$^\dag$ & 6.43 & 57.0 & 91.6 & 3.20 & 5.25\\
\bottomrule
\multicolumn{5}{r}{$^*$\footnotesize{Monocular depth method.} $^\dag$ \footnotesize{Method uses monocular depth.} }
\end{tabular}
}
\end{center}
\caption{Run-time, memory and complexity evaluation of all methods with a single 350$\times$630 RGB image on an NVIDIA GeForce RTX 3080 GPU. The choice of resolution is dictated by the baseline methods which run out of memory for larger images on the specified hardware. GMACS are Giga Multiply-Accumulate ops./second.}
\label{table:efficiency-comparison}

\end{table}

\subsection{Implementation}
Our method is implemented in PyTorch and trained on eight Nvidia Tesla V100 GPUs. We use 256$\times$384 images and accumulate gradients across eight mini-batches of 16 samples each. For both networks $\Theta(\cdot)$ and $\Psi(\cdot)$, we use the Adam optimizer with a learning rate of $1\times10^{-3}$ and a cosine annealing schedule with restarts every 200 epochs. A vectorized implementation of the differentiable TMPI renderer (Algorithm~\ref{algo:render-tmpi}) runs at $
\sim$100ms~/~mini-batch allowing efficient parallel computation.

\subsection{Baselines}
We compare our approach to four state-of-the-art single-view 3D photography methods based on multiplane images: Tucker~\etal~\cite{tucker2020} (SVMPI), Li~\etal~\cite{li2020} (VMPI), Li~\etal~\cite{li2021} (MINE) and Han~\etal~\cite{han2022} (AdaMPI). Like us, VMPI and AdaMPI use a monocular estimator to recover depth as the first step of their pipeline. The input to all methods is the same, however --- a single unconstrained RGB image --- and so we evaluate them as end-to-end 3D photography approaches. Nonetheless, we do note the additional depth estimation step when evaluating computational and memory performance (Table~\ref{table:efficiency-comparison}).  We use $N=32$ planes for all baselines except VMPI, which is designed for $N=8$.

We do not compare to single-view methods based on neural radiance fields~\cite{yu2021} or the recent work of Nicklaus~\etal~\cite{niklaus2019} and Shih~\etal~\cite{shih20} as these are too computationally intensive for our intended use cases on mobile and VR devices. While recent work on NeRFs has demonstrated impressive rendering speeds, training remains expensive, and further, requires a large number of input views and a static scene. Jampani~\etal's~\cite{jampani2021} method, though not MPI-based, is related. But the authors have not released their code.

\subsection{Testing Datasets}
\begin{figure*}[h!]
   \includegraphics[width=1.0\linewidth, trim={0cm, 9cm, 0cm, 0cm}, clip]{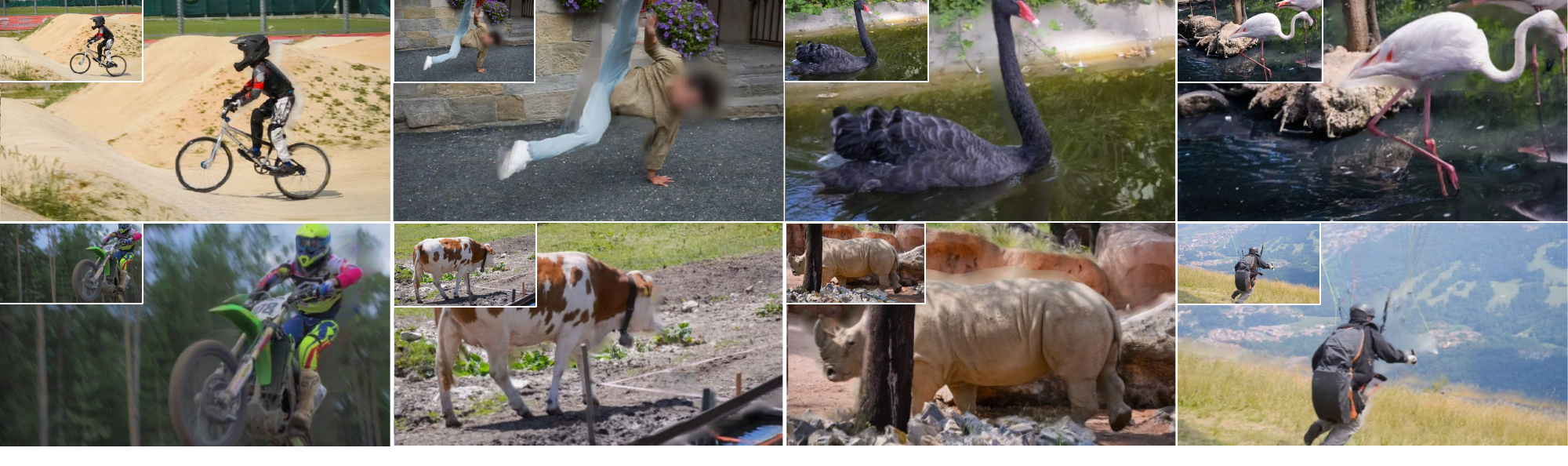}
\caption{View synthesis results of our method on the HD (1080$\times$1920) Davis~\cite{ponttuset2017} dataset. Original views are inset. }
\label{fig:hires-results}
\vspace{-3mm}
\end{figure*}

We test all methods on the \textit{Spaces}~\cite{flynn2019}, and \textit{Tanks and Temples}~\cite{knapitsch2017} datasets. \textit{Spaces} consists of 100 indoor and outdoor scenes captured using a purpose-built, 16-camera rig. For \textit{Tanks and Temples} we use the \emph{Intermediate} split which has uniformly sampled frames from eight high-resolution videos of more challenging outdoor environments. We compute camera poses and depth maps for all scenes using COLMAP~\cite{schoenberger2016mvs, schoenberger2016sfm}. The depth maps are required to resolve the scale ambiguity of monocular depth for correct reprojection to target views. We randomly select 1000 source views from each dataset scaled to 350$\times$630 and use the next image in capture sequence as the target for view synthesis. This choice of resolution is dictated by the baseline methods which run out of memory for larger images on the specified hardware. We present high resolution results for our method on the Davis dataset~\cite{ponttuset2017} in Figure~\ref{fig:hires-results}.

\subsection{Evaluation Metrics}
We evaluate the rendered views quantitatively on four metrics: Peak-Signal to Noise Ratio (PSNR), Structural Similarity~\cite{wang2004} (SSIM), Learned Perceptual Image Patch Similarity~\cite{zhang2018} (LPIPS) with a VGG-16 backbone, and the mean absolute error (L1). Following previous work~\cite{han2022, tucker2020, li2020}, we crop 15\% of the image around the edge to account for disocclusions. Further, if more than 15\% of the remaining pixels are blank in a view synthesized by any method, we discard the result across all baselines.

\begin{figure*}[h!]
   \includegraphics[width=1.0\linewidth, trim={0cm, 0.15cm, 12cm, .25cm}, clip]{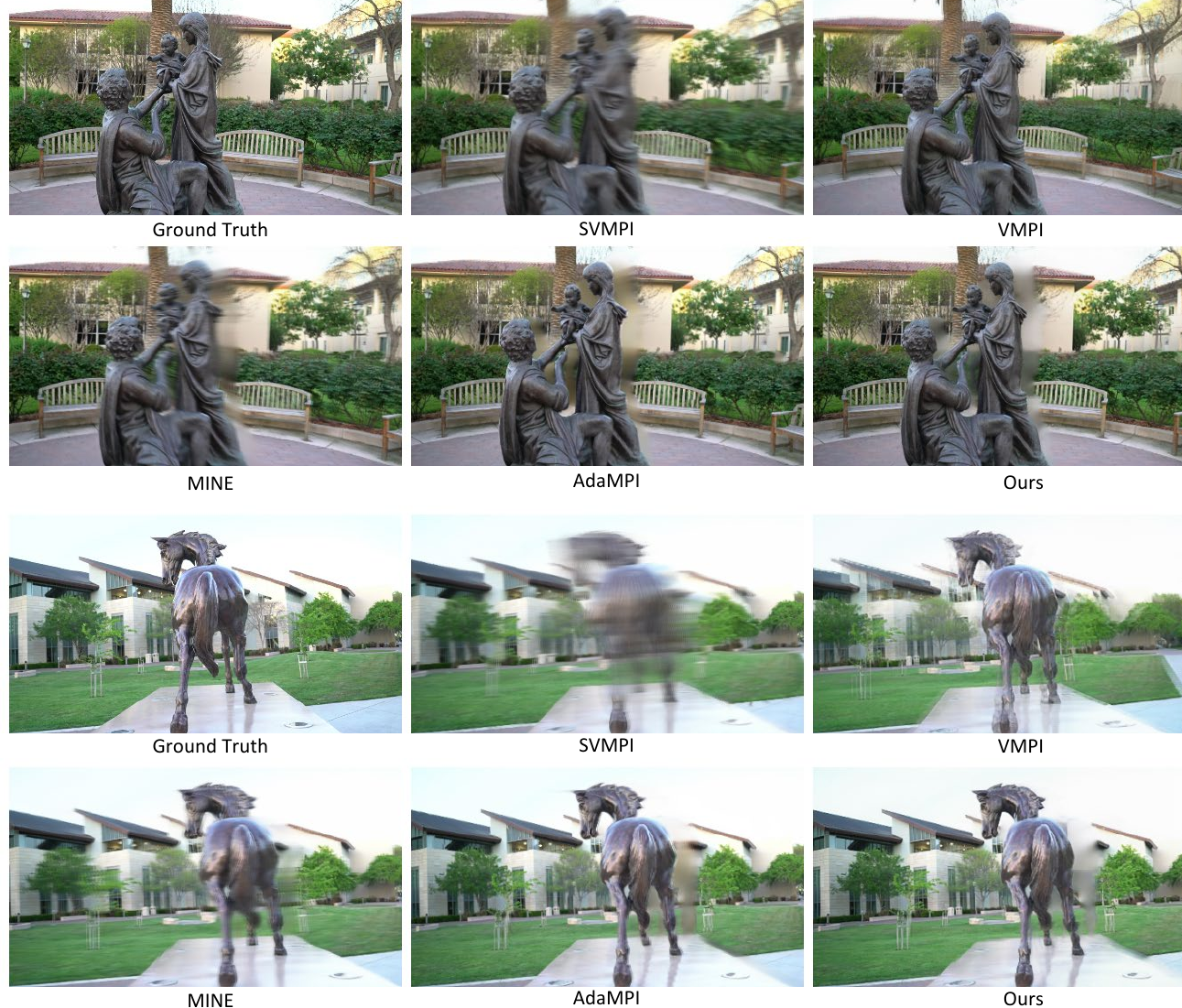}
\caption{Comparing the novel view synthesis results of the baseline methods and our approach on the \textit{Tanks and Temples} dataset. Our results are better than SVMPI, VMPI and MINE, and comparable to AdaMPI while using far fewer depth planes. }
\label{fig:results}
\end{figure*}

\begin{figure}[h!]
   \includegraphics[width=1.0\linewidth, trim={0cm, 0cm, 17cm, 0cm}, clip]{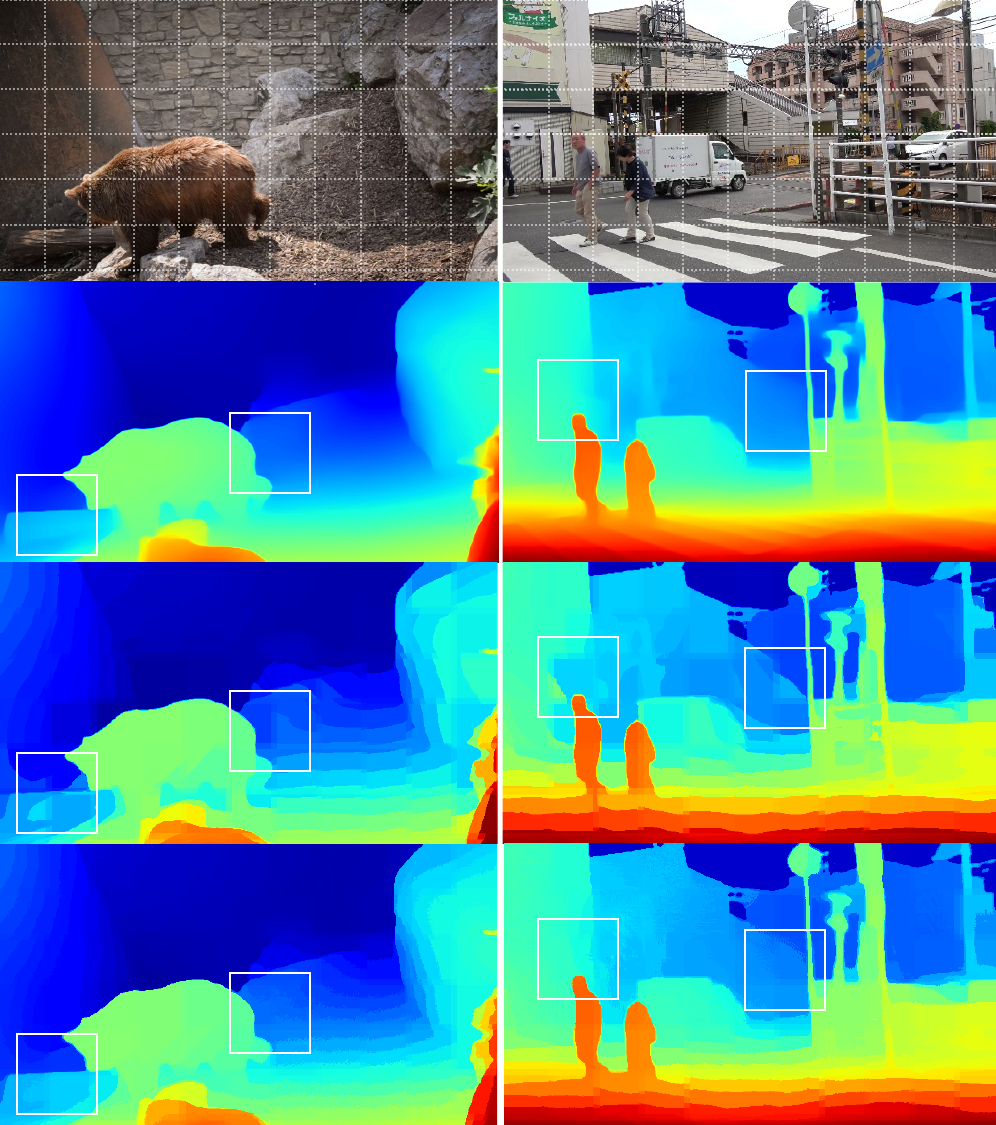}
\caption{\textbf{Top to bottom:} Input image (Davis dataset); inverse monocular depth from DPT~\cite{ranftl2021}; naive discretization by spacing four planes linearly in the inverse-depth range of each tile; our weighted clustering-based discretization with four planes. Our representation shows finer variation on receding surfaces and suffers fewer tiling artifacts. }
\label{fig:results-depth}
\vspace{-2mm}
\end{figure}
\setlength{\tabcolsep}{4.5pt}
\setlength{\tabcolsep}{4pt}
\begin{table}
\begin{center}
\begin{tabular}{lccc}
\toprule
\multicolumn{4}{c}{Spaces Dataset} \\
\midrule
Variant & MAE $\downarrow$ & MSE $\downarrow$ & Q25 $\downarrow$ \\
\midrule
Vanilla $k$-means & 35.3 & 2.00 & 14.0\\
Linear plane spacing & 48.7 & 3.80 & 19.3 \\
Ours & {27.1} & {1.20} & {10.6} \\

\midrule

\multicolumn{4}{c}{Tanks\& Temples Dataset} \\
\midrule
Variant & MAE $\downarrow$ & MSE $\downarrow$ & Q25 $\downarrow$ \\
\midrule
Vanilla $k$-means & 35.7 & 2.00 & 14.2\\
Linear plane spacing & 46.0 & 3.40 & 18.1 \\
Ours & {27.0} & {1.20} & {10.6} \\

\bottomrule

\end{tabular}
\end{center}
\caption{Evaluating the reconstruction error of different depth discretization approaches. All values are $\times10^{-3}$.}
\vspace{-3mm}
\label{table:ablation}
\end{table}

\subsection{Results}
\begin{figure*}[h!]
    \centering
   \includegraphics[width=0.9\linewidth, trim={0cm, 0cm, 20cm, .25cm}, clip]{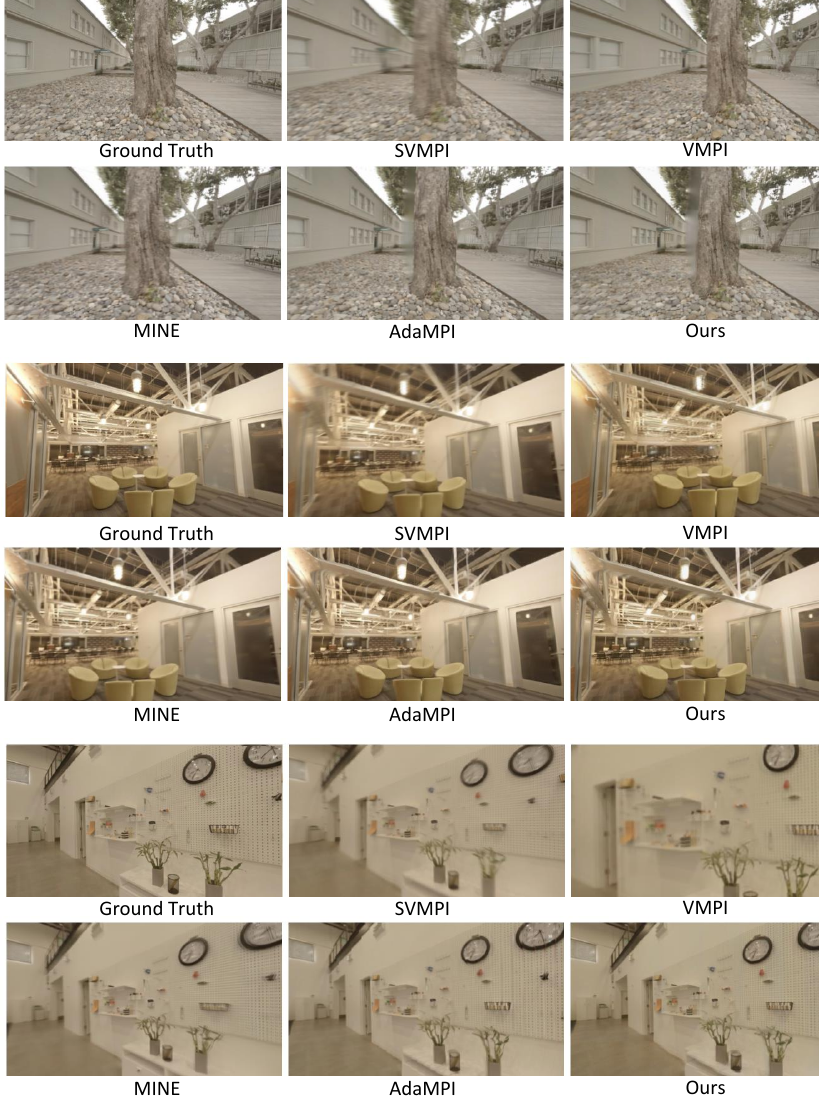}
\caption{Comparing the novel view synthesis results of the baseline methods and our approach on the \textit{Spaces dataset}. }
\label{fig:results-spaces}
\end{figure*}

Table~\ref{table:mpi-results} presents quantitative evaluation of all methods on the two test sets. Qualitative results are shown in Figure~\ref{fig:results} for \textit{Tanks and Temples} and Figure~\ref{fig:results-spaces} for \textit{Spaces}. Our approach uses $n=4$ depth planes with a tile size of $h=64$.  The competitive performance of our method despite having much fewer planes per tile can be attributed to the adaptive placement of depth planes in each tile which allows it to effectively cover a larger depth range than the monolithic MPIs of the baselines, 
Memory and computational performance is evaluated in Table~\ref{table:efficiency-comparison}. VMPI, AdaMPI and our method have the additional overhead of monocular depth estimation using Ranftl~\etal's~\cite{ranftl2021} DPT. Our approach has lower runtime,  peak memory and space requirements than AdaMPI while achieving similar quality results. 

We evaluate our plane placement strategy in Table~\ref{table:ablation}. We measure the reconstruction quality of the discretized depth map defined by the planes in each tile on a monocular depth input that is perturbed by a small amount of Gaussian noise ($\lambda=0$, $\sigma^2=1\times10^{-3}$). We compare our approach to naive linear spacing of depth planes in inverse disparity space, and to vanilla unweighted $k$-means. Our method is robust to outliers and yields a much better reconstruction. Figure~\ref{fig:results-depth} provides qualitative comparison with linear spacing on samples from the Davis dataset~\cite{ponttuset2017}.

\begin{figure}[t]
    \includegraphics[width=0.95\columnwidth, trim={0cm, 7.0cm, 17.5cm, 0cm}, clip]{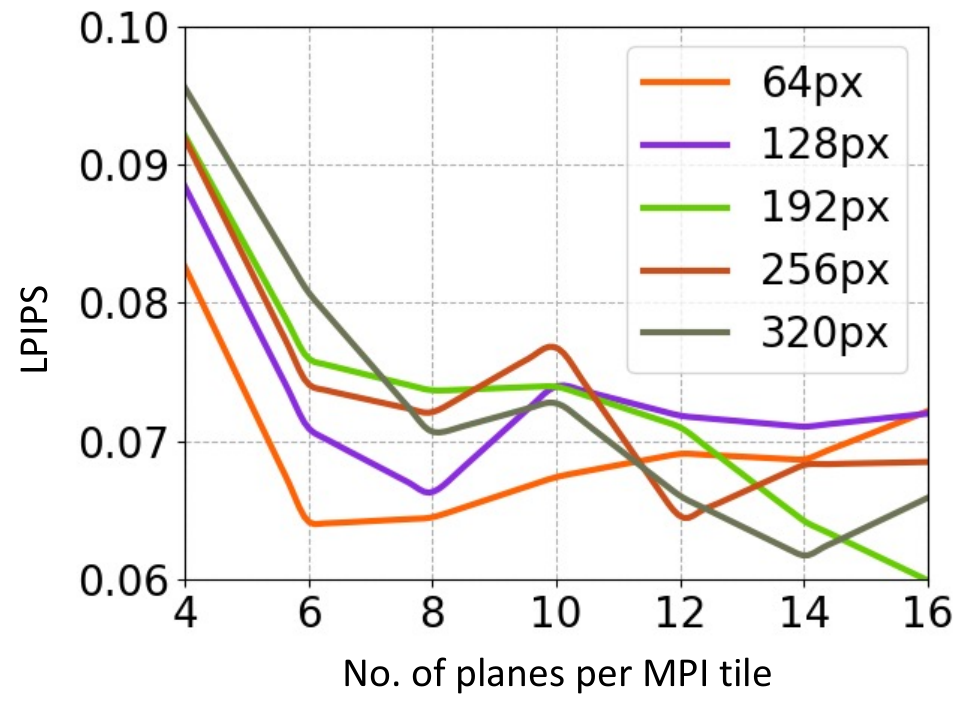}
    \caption{Evaluating the effect of tile size and number of depth planes on view synthesis quality. In general, the results uphold the intuition that small tiles and a large number of planes improve quality. The trend, however, is less clear when the number of planes $>8$, indicating that the model may be unable to handle redundant geometry effectively. }
    \label{fig:lpips-planes-tiles}
\end{figure}

Figure~\ref{fig:lpips-planes-tiles} evaluates the effect of tile size and number of planes on the quality of view synthesis.   In general, the results uphold the intuition that small tiles and a large number of planes improve quality. This trend is less clear, however, for number of planes $>8$. This would seem to support the observation of Khakhulin~\etal~\cite{khakhulin2022} and Hu~\etal~\cite{hu2021} that reductive models are unable to handle redundant geometry effectively. Moreover, the  model's complexity increases proportionately with the number of depth planes, leading to slower convergence for the same number of training steps.

\section{Limitations}
\label{sec:limitations}
\begin{figure}[t]
\centering
    \includegraphics[width=1.0\columnwidth, trim={0cm, 14cm, 18cm, 0cm}, clip]{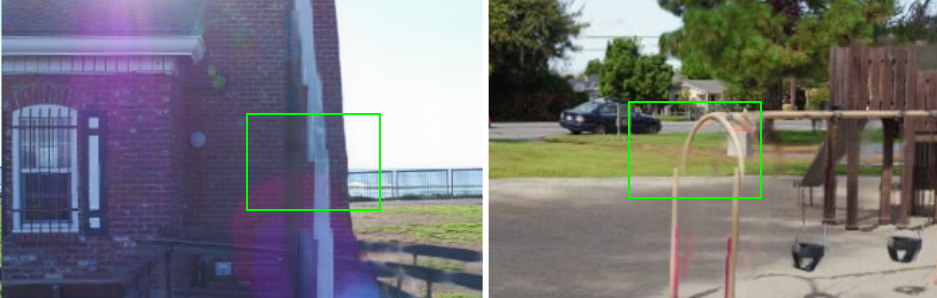}
    \caption{Failure cases of our approach: in some cases, fine features are inconsistently reconstructed across tiles.}
    \label{fig:limitations}
\end{figure}

As previously observed, a tiled multiplane image cannot exploit the elegant warping and compositing equations of a traditional MPI for differentiable rendering during training. Furthermore, in some cases our method fails to reconstruct thin features consistently across tiles (Figure~\ref{fig:limitations}). 

\section{Conclusion}
\label{sec:conclusion}
We present a method for estimating \emph{tiled multiplane images} from a single RGB input for 3D photography. This includes a novel approach to adaptively spacing a small number of depth planes within an MPI tile to better represent local features. Our method is lightweight, and points a path to realizing novel view synthesis on mobile and VR devices. 

{\small
\bibliographystyle{ieee_fullname}
\bibliography{main}
}

\end{document}